\documentclass[sigconf]{acmart}
\usepackage{autobreak}

\usepackage{multirow}
\usepackage{geometry}
\usepackage{algorithm}
\usepackage{algorithmic}
\usepackage{amsthm}
\newtheorem{theorem}{Theorem}
\newtheorem{lemma}{Lemma}
\newtheorem{defi}{Definition}

\usepackage{graphicx}
\usepackage{amsmath}
\AtBeginDocument{%
  \providecommand\BibTeX{{%
    \normalfont B\kern-0.5em{\scshape i\kern-0.25em b}\kern-0.8em\TeX}}}

\setcopyright{acmcopyright}
\copyrightyear{2018}
\acmYear{2018}
\acmDOI{10.1145/1122445.1122456}

\acmConference[Woodstock '18]{Woodstock '18: ACM Symposium on Neural
  Gaze Detection}{June 03--05, 2018}{Woodstock, NY}
\acmBooktitle{Woodstock '18: ACM Symposium on Neural Gaze Detection,
  June 03--05, 2018, Woodstock, NY}
\acmPrice{15.00}
\acmISBN{978-1-4503-XXXX-X/18/06}

\begin{document}

\title{A General Traffic Shaping Protocol in E-Commerce}

\author{Chenlin Shen}
\affiliation{%
  \institution{Alibaba Group}
  \country{Hangzhou, China}
}

\author{Guangda Huzhang}
\email{guangda.hzgd@alibaba-inc.com}
\affiliation{%
  \institution{Alibaba Group}
  \country{Hangzhou, China}
}

\author{Yuhang Zhou}
\affiliation{%
  \institution{Alibaba Group}
  \country{Hangzhou, China}
}

\author{Chen Liang}
\affiliation{%
  \institution{Alibaba Group}
  \country{Hangzhou, China}
}

\author{Qing Da}
\affiliation{%
  \institution{Alibaba Group}
  \country{Hangzhou, China}
}


\begin{abstract}
  To approach different business objectives, online traffic shaping algorithms aim at improving exposures of a target set of items, such as boosting the growth of new commodities. Generally, these algorithms assume that the utility of each user-item pair can be accessed via a well-trained conversion rate prediction model. However, for real E-Commerce platforms, there are unavoidable factors preventing us from learning such an accurate model. In order to break the heavy dependence on accurate inputs of the utility, we propose a general online traffic shaping protocol for online E-Commerce applications. In our framework, we approximate the function mapping the bonus scores, which generally are the only method to influence the ranking result in the traffic shaping problem, to the numbers of exposures and purchases. Concretely, we approximate the above function by a class of the piece-wise linear function constructed on the convex hull of the explored data points. Moreover, we reformulate the online traffic shaping problem as linear programming where these piece-wise linear functions are embedded into both the objective and constraints. Our algorithm can straightforwardly optimize the linear programming in the prime space, and its solution can be simply applied by a stochastic strategy to fulfill the optimized objective and the constraints in expectation. Finally, the online A/B test shows our proposed algorithm steadily outperforms the previous industrial level traffic shaping algorithm.
%   The main difference between our proposed LP and that used in existing online matching algorithms is that the optimization variables of our proposed LP are bonus scores, thus we directly optimize the LP in the primal space to obtain optimal bonus scores. Then we apply the optimal bonus scores by a randomized combination of explored bonus scores, which is shown to be better than a deterministic protocol. 
\end{abstract}

\begin{CCSXML}
<ccs2012>
<concept>
<concept_id>10003752.10010070</concept_id>
<concept_desc>Theory of computation~Theory and algorithms for application domains</concept_desc>
<concept_significance>500</concept_significance>
</concept>
<concept>
<concept_id>10010405.10003550.10003555</concept_id>
<concept_desc>Applied computing~Online shopping</concept_desc>
<concept_significance>500</concept_significance>
</concept>
</ccs2012>
\end{CCSXML}

\ccsdesc[500]{Theory of computation~Theory and algorithms for application domains}
\ccsdesc[500]{Applied computing~Online shopping}
\keywords{Traffic shaping, black-box approximation, linear programming}


\maketitle

\section{Introduction}
Most E-Commerce algorithms aim to improve transaction efficiency by showing personalized items according to the interests of users, which provides the basic power for the growth of the platform. However, transaction efficiency is not the only business concern in many cases. For example, the delivery timeliness is a vital feature affecting the satisfaction of users, since users wish to receive their package as soon as possible. Therefore, for the sake of optimizing the satisfaction of users, retail platforms will boost certain exposure for the items with better delivery timeliness. The demand for such quantitative exposure boosting is fundamental in E-Commerce operations and can be formulated as an online traffic shaping problem. The goal is to maximize the cumulative purchase number while satisfying the constraints of the lowest exposure number.

Traditional online traffic shaping algorithms assume that the utility of each user-item pair can be accessed via an oracle model. In other words, there exists a model that serves as an oracle to accurately predict the probability of purchase when showing an item to a user. With the assistance of the probabilities, the traffic shaping problem can be further formulated as a linear program (LP).
Moreover, by adopting the stochastic user arrival model, most existing algorithms are based on the primal-dual framework, where dual optimal prices are learned by solving a fractional LP with the probabilities of revealed users and are used for subsequent assignments.
However, there are unavoidable factors that prevent us from learning an accurate model to predict the probabilities, such as the cold-start problem, class-imbalance data set, and heavy noises in behavior patterns. 
% Moreover, most ranking models are trained with rank loss, where the goal is to obtain a correct permutation rather than learning the accurate probabilities. In this situation, the model-output scores are used as the keywords of sorting items and are often far from the real probabilities of purchase. 
Therefore, the traditional online traffic shaping algorithms can be ineffective in practice, leading to either non-optimal purchase numbers or severe violation of the exposure constraints.

To break the heavy dependence on accurate probabilities, we propose a novel and general online traffic shaping protocol for E-Commerce platforms. Instead of trying to directly learn the dual optimal prices with ranking scores, we treat the ranking model as a component of the whole environment. In this environment, we introduce functions that map the dual prices to the numbers of exposure and purchase on each user group. For the ease of intuitive understanding, we rename the dual price as \textit{bonus score} that can be added to the model-output scores in the rest of this paper. Since the bonus score is a continuous variable, we explore the function values on several points and approximate the function with a piece-wise linear function constructed on the convex hull of the explored points. Moreover, we reformulate the online traffic shaping problem as an LP where the piece-wise linear functions are embedded into both the objective and constraints. The main difference between our proposed LP and that used in existing online matching algorithms is that the optimization variables of our proposed LP are bonus scores, thus we directly optimize the LP in the primal space to obtain optimal bonus scores. Then we apply the optimal bonus scores by a randomized combination of explored bonus scores, which is shown to be better than a deterministic protocol. 

The main contributions of this paper are summarized as follows:
\begin{itemize}
\item We propose a general online traffic shaping protocol for E-Commerce platforms to break the heavy dependence on accurate probabilities in online traffic shaping problems.
\item We proof that our proposed protocol achieves the optimal solution in the stationary environments.
\item The experiments on real applications exhibit the superiority of our approach.
\end{itemize}

\section{Related Works}
A close research topic of traffic shaping is online matching~\cite{DBLP:journals/fttcs/Mehta13}. Generally, online matching algorithms focus on optimizing the competitive ratio given a user arriving model, e.g. \cite{fahrbach2020edge,agrawal2014dynamic,DBLP:journals/fttcs/BuchbinderN09,feldman2010online}. Some recent works study the efficiency on converge~\cite{agrawal2014fast,agrawal2016efficient,li2020simple}, extension versions for more general setting~\cite{aggarwal2011online,kesselheim2013optimal,huzhang2017online}, and more complicated user models~\cite{esfandiari2015online,zhou2019robust}. Different from online matching, online convex optimization involves learning frameworks and has been studied in theory and practice~\cite{hazan2019introduction}. Recent online convex optimization algorithms study dynamic regret and can adapt to both stationary and dynamic environments~\cite{zinkevich2003online,besbes2015non,zhang2018adaptive,zhao2020dynamic}. Compared to them, our study focuses on real-world scenarios, and aims to build the complete process that uses the raw statistics data to improve the online revenue which is a complicated black-box function.

\section{Preliminary}
Our online platform has $n$ disjoint \emph{user groups} $U = \{U_1, U_2, ..., U_n\}$ and target items set $S = \{S_1, S_2, ..., S_m\}$. User groups are partitioned by business experts and the users in the same group are likely to have similar behaviors. In our online platform, items displayed to users are ranked by their \emph{ranking scores}, and we increase the exposure of a target items set by adding the a
\emph{bonus score} to these items. The bonus scores for items in the same target items set should be the same, and they can be different given different users. Practically, there may be items which belong to two target items set, but the amount of these items can be ignored. Therefore, we assume target items sets are also disjoint.

For a traffic shaping task, our first objective is to satisfy \emph{the lowest exposures requirement} of each target set $S_j$. Concretely, the total exposure of items in target set $S_j$ should be at least $r_j$. Our second objective is to \emph{minimize the loss of clicks} caused by the above exposures requirement. Formally,  
\begin{equation}
\label{obj}
\begin{aligned}
maximize &\sum_{1\leq i \leq n, 1 \leq j \leq m} g_{ij}(x_{ij})  \\
s.t. \quad &\sum_{1\leq i \leq n} f_{ij}(x_{ij}) \geq r_{j}, \forall j \in [1, m] \\
\end{aligned}
\end{equation}
%Modern ranking systems generally sort items in a score-based style: each item gets a score from ranking systems as the keyword of sorting. To be compatible with them, we assign a bonus score to the items we want to lift their exposures. 
The \emph{non-decreasing} function $f_{ij}(x)$ represent the times that items in the $j$-th set will be exposed for buyers in the $i$-th group when we assign bonus score of $j$-th set to $x$ in expectation (to dispose the uncertainty of an online environment). 
The \emph{non-increasing} function $g_{ij}(x)$ represent the times of \emph{total clicks} (for all items) under the same setting.
%Here $f_{ij}(x)$ is non-decreasing and $g_{ij}(x)$ is non-increasing when the bonus score becomes larger.

% Let \emph{monotonic} functions $f_{ij}(x)$ and $g_{ij}(x)$ represent the times that items in the $j$-th set will be exposed and clicked for buyers in the $i$-th group when we assign bonus score of $j$-th set to $x$. We want to maximize the following programming:

\section{Methodology}
We can observe that Equation~\ref{obj} is a \emph{non-linear} programming and it is hard to solve it straightforwardly. The high level idea of our solution is to estimate and break $f$ and $g$ into the sum of several piece-wise linear functions. Without loss of generality, we assume the bonus score is chosen from $0$ to $1$. We partition the interval into $k+1$ sub-intervals and examine a fixed set of bonus score value $T = (0, 1/k, 2/k, ..., 1)$. We estimate the ground-truth value $(f(x),g(x))$ for $x_{lim}\in T$, denoted as $(\bar{f}(x_{lim}),\bar{g}(x_{lim}))$. Then, a stochastic allocation can be proved to optimize the performance. We demonstrate the process by pseudo code in Algorithm~\ref{alg:main}. %We list the involved notations as follows.

%Suppose we observe several independent records for each possible bonus score in $T$. We want to estimate the ground-truth expected value $(f(x),g(x))$ by these observations. Let $(\bar{f}(x_{lim}),\bar{g}(x_{lim}))$ be the approximated expected value according to the observation, where $x_{lim}\in T$. Our objective is to find a protocol  for each possible $x$, which makes the number of exposures close to $f(x)$ and the number of clicks greater than $g(x)$.

\begin{algorithm}
\caption{AE Traffic Shaping}
\label{alg:main}
\begin{algorithmic}
\STATE {\textbf{Input: }  Oracle functions of exposures $(f_{ij})$ and clicks $(g_{ij})$, {the lowest exposures requirement} $r_j$.}
\STATE {\textbf{Output: } A stochastic bonus score assignment for $\{x_{ij}\}$.}
\STATE {}
\STATE {Constraints set $C \gets \emptyset$}
\STATE {Optimization variables set $X \gets \emptyset$}
\STATE {Objective function $O\gets 0$}
\STATE {Split points $T \gets (0, 1/k, 2/k, ..., 1)$}
\FOR {$i \in [n], j \in [m]$}
\STATE {Detect $f_{ij}(x_{\text{lim}})$ and $g_{ij}(x_{\text{lim}})$ for $x_{\text{lim}} \in T$, denoted as $\bar{f}_{ij}(x_{lim})$ and $\bar{g}_{ij}(x_{lim})$}% // expensive interactions with oracles}
\STATE {Solve \emph{the outer convex curve} $L$ of points $(\bar{f}_{ij}(x_{lim}), \bar{g}_{ij}(x_{lim}))$ by the Graham's scan algorithm // refer Definition~\ref{def:ocl}}
\STATE {Sort points of $L$ in increasing order of $x$}
\FOR {$k = 2$ to size of L}
    \STATE {$(\bar{f}_{ij}(x_{k}), \bar{g}_{ij}(x_{k})) \gets $ $k$-th element of L}
    \STATE {$(\bar{f}_{ij}(x_{k-1}), \bar{g}_{ij}(x_{k-1})) \gets $ $(k-1)$-th element of L}
    \STATE{$X \gets X \cup \{x_{ijk}\}$}
    \STATE {$\text{mixedBy}(x_{ijk}) \gets (x_{k-1}, x_k)$}
    \STATE{$C \gets C \cup \{0\leq x_{ijk} \leq \bar{f}_{ij}(x_{k}) - \bar{f}_{ij}(x_{k-1})\}$}
    \STATE{$O \gets O + x_{ijk} \cdot \frac{\bar{g}_{ij}(x_{k-1}) - \bar{g}_{ij}(x_{k})}{\bar{f}_{ij}(x_{k}) - \bar{f}_{ij}(x_{k-1})}$ }
\ENDFOR

\ENDFOR
\STATE {Solve LP variables $X$ that optimizes $O$ under constraints $C$}
\STATE {Initialize stochastic bonus score assignment $A$}

\FOR {$i \in [n], j \in [m]$}
\STATE {Find maximal $t$ that $x_{ijt} > 0$}
\STATE {$x_{\text{left}}, x_{\text{right}} \gets \text{mixedBy}(x_{ijt})$}
\STATE {Add the following rule to $A$: let $x_{ij}=x_{\text{left}}$ with probability of $1-\frac{x_{ijt}}{x_{\text{right}}  - x_{\text{left}}}$, and $x_{ij}=x_{\text{right}}$ with probability of $\frac{x_{ijt}}{x_{\text{right}} - x_{\text{left}}}$}
\ENDFOR
\RETURN {$A$}
\end{algorithmic}
\end{algorithm}

% We notice if unknown function $f$ and $g$ are linear, then the optimization task becomes easy. Therefore, our solution is to approximate these non-linear functions by piece-wise linear functions. Without loss of generality, we assume the bonus score can be chosen from $0$ to $1$. We partition the interval into $k+1$ sub-intervals and examine a fixed set of bonus score value $T = (0, 1/k, 2/k, ..., 1)$. 

\subsection{Proof of Main Theorem}
% In our mechanism, as well as any score-based ranking systems, we prove any $f$ can be well approximated by a reasonably large parameter $k$ for $T$ when $\bar{f}, \bar{g}$ are accurate.
We purify our theoretic contribution in the following theorem. Practically, we will choose a reasonably large parameter $k$ and go through the steps in Algorithm~\ref{alg:main}.
\begin{theorem}
\label{protocol}
Assume $|\bar{f}_{ij}(x_{lim})- f_{ij}(x_{lim})|<\epsilon_1$ and $|\bar{g}_{ij}(x_{lim}) - g_{ij}(x_{lim})|<\epsilon_2$ for $x_{lim}\in T$ given monotonic and continuous function $f_{ij}$ and $g_{ij}$. 
When $k$ goes to infinity, there is a polynomial-time protocol that produces the expected number of exposures $\hat{f}_{j}$ for each target set $j$ and the global expected number of clicks $\hat{g}$, 
satisfying $\mathbb{E}[\hat{f}_{j}] + n\epsilon_1 \geq r_j$  and $E[\hat{g}] + nm\epsilon_2  \geq g^*$ where $g*$ is the global optimality on clicks.
% satisfying $\mathbb{E}[\hat{f}_{j}] + n\epsilon_1 \geq r_j$  and $E[\hat{g}] + nm\epsilon_1 \cdot \sup(|g'|)  + nm\epsilon_2  \geq g^*$ where $g*$ is the global optimality.
%Given intervals number $k$, if $Pr[|\bar{f}(x) - f(x)| < \epsilon] \geq 1 - \delta$, we can build a protocol that the expected number of exposures $\hat{f}(x)$ and  the expected number of clicks  $\hat{g}(x)$  satisfy $Pr[|\hat{f}(x) - f(x)| \leq \epsilon + \frac{1}{2k} ] \geq 1-2\delta$ and $Pr[\hat{g}(x) \geq g(x) - \epsilon - \frac{1}{2k} ] \geq 1-2\delta$ for each x.
\end{theorem}

We first introduce a series of lemmas to prove the theorem. The first lemma allows us to estimate the effect of an arbitrary bonus score by a stochastic combination of bonus scores that already have been observed.
\begin{lemma}
\label{line}
Given values of $f$ and $g$ in two points $x_0$ and $x_1$, a stochastic strategy can obtain exposures $\lambda f(x_0) + (1- \lambda ) f(x_1)$ and clicks $\lambda g(x_0) + (1- \lambda ) g(x_1)$ for any $\lambda \in [0, 1]$ in the sense of expectation.
%for each $0 \leq r \leq 1$.  
\end{lemma}
\begin{proof}
We can adopt the bonus score $x_0$ with the probability of $\lambda$ and the bonus score $x_1$ with the probability of $1-\lambda$. It is easy to see the desired result.
\end{proof}

%Another issue is that the traditional piece-wise linear approximation is still non-linear. 
The following lemma introduces a special class of piece-wise linear functions that can be written without conditions on pieces. 
\begin{lemma}
\label{convex}
Given a non-increasing, continuous, and piece-wise linear function $g$ separated by $k$ points $\{s_1, s_2, ..., s_k\}$,
\begin{equation}
g(x) = \{k_i (x - s_i) + b_i, s_i \leq x \leq s_{i+1} | i \in [1, k-1]\}.
\label{eq:ofmaxsum}
\end{equation}

The above formula can be written as 
\begin{equation}
g(\sum x_i) = b_1 + \sum_{i=1}^{k-1} k_i x_i \text{ if }  (x_i > 0 \implies x_j = s_{j+1} - s_{j}, \forall j<i)
\label{eq:fsum}
\end{equation}

The condition $\{x_j = s_{j+1} - s_{j}, \forall j<i\}$ implies that all previous $x_j$ are maximized. If $k_i \geq k_{i+1}$ for each $i$, then 
%\begin{equation}

%\frac{y_{i-1} - y_{i-2}}{x_{i-1} - x_{i-2}} \leq \frac{y_i - y_{i-1}}{x_i - x_{i-1}}, \foreach i \in [3, k]
%\end{equation}

\begin{equation}
g(x) = b_1 + \max_{x_i \in [0, s_{i+1}-s_i], \sum x_i=x} [\sum_{i=1}^{k-1} k_i x_i]
\label{eq:fmaxsum}
\end{equation}
\end{lemma}

\begin{proof}
By the definition, we can see $g(s_1) = b_1$. Consider the process of value changing of $g$ when $x$ moves from $s_1$ to $s_k$. Its gradient is $k_1$ at the beginning. When the first interval is exhausted (i.e. $x\geq s_1$), the speed of value changing becomes $k_2$.
 Equation~\ref{eq:fsum} limits the added portion of $x$  must be selected from left to right so it is equivalent to the original definition of $g(x)$.
 
 Equation~\ref{eq:fmaxsum} demonstrates a different process where added portion of $x$ is not constrained to from left to right. Because of the increasing property of $\{k_i\}$, maximizing $\sum k_i x_i$ implies that $x$ should after fill the left interval before go right. Therefore, Equation~\ref{eq:fsum} and \ref{eq:fmaxsum} is equivalent with increasing $\{k_i\}$.

% Suppose we have $x$ units of resources and we need to distribute them to $k-1$ people. The $i$-th person produces $k_i x_i$ revenue if we give $i$-th person $x_i$ units of resources and he can get at most $s_{i+1}-s_i$ units of resources. Then maximizing the revenue can be written as the second term in the right of equation~ref{convex}. Because $f$ is convex, allocating the resource to the person with less index always produces the best efficiency. Therefore, this maximizing task is the same process as the value-changing process of $f$.
\end{proof}
%The above lemma is applicable for $g$ when $\{k_i\}$ is increasing. 
%The maximization operation will be absorbed when we rewrite $g$, but $f$ appears in the constraints and cannot follow the same trick. 
The next lemma introduces a surrogate objective $g_c$ can promise that $\{k_i\}$ is increasing and objective value $g_c(x)$ is not worse than $g(x)$ for each $x$.

\begin{defi}
The outer convex curve for a set of points $\{(x_i, y_i)\}$ is obtain by eliminating $(x, y)$ if there exist $(x', y')$ and $(x'', y'')$ s.t. $x'\leq x \leq x''$ and $(y - y') \cdot (x''-x') <  {(x-x')\cdot(y''-y')}$. The outer convex curve is the upper-half of a convex hull and can be computed by the Graham's scan algorithm.
\label{def:ocl}
\end{defi}
\begin{lemma}
\label{convexhull}
Given a monotone, continuous, piece-wise linear function  $g$ separated by $k$ points $\{s_1, s_2, ..., s_k\}$, the new piece-wise linear function $f^c$ that describes the outer convex curve for $\{(s^{'}_k, f(s^{'}_k))\}$ satisfies $f^c(x) \geq f(x)$ for an arbitrary $x$.
\end{lemma}

\begin{proof}
For $s_i \leq x \leq s_{i+1}$ and both $(s_i, f(s_i))$ and $(s_{i+1}, f(s_{i+1}))$ are in the outer convex curve, $f^c(x) = f(x)$. Otherwise, without loss of generality, $(s_i, f(s_i))$ can be eliminated by two points $(x', f(x'))$ and $(x'', f(x''))$ in the outer convex curve, which satisfies $x' \leq s_i \leq x''$and
\begin{equation}
\begin{aligned}
    &(f(s_i) - f(x')) \cdot (x''-x')<{(s_i-x')\cdot(f(x'')-f(x'))} \\
    \implies &f(s_i) < f(x') + \frac{(s_i-x')\cdot(y''-y')}{x''-x'}=f^c(s_i)
\end{aligned}
\end{equation}
Note that the singular case $x' = s_i = x''$ is solved naturally. The above result shows $f^c\geq f$ at least on separated points, which can extends to arbitrary points for piece-wise linear functions.
% The result can be figured out by contradiction. If there is a $x$ satisfying  $f^c(x) < f(x)$ and $s_i \leq x \leq s_{i+1}$, then either $f^c(s_i) < f(s_i)$ or $f^c(s_{i+1}) < f(s_{i+1})$, implying that
% \begin{equation}
    
% \end{equation}

% but both of them imply that $f_c$ does not contain all points in the convex hull.
\end{proof}

%We then go through the proof of our main theorem.

\begin{proof}
\textbf{(of Theorem~\ref{protocol})} 
Let $OBJ$ the optimized value of objective in Equation~\ref{obj}. We first transform Equation~\ref{obj} by replacing $f$ and $g$ with the summation of linear functions in Equation~\ref{eq:fsum}. We use $\bar{f}_{ijt}$ and $\bar{g}_{ijt}$ to represent the gradient of each linear function ($k_i$ in Equation~\ref{eq:fsum}), i.e. $\bar{f}_{ij}(\sum x_t)= \bar{f}_{ij}(0)+\sum_{t=1}^{k-1}\bar{f}_{ijt} x_t$. After replacing $f$ and $g$ with $\bar{f}$ and $\bar{g}$, we have
% $\bar{f}_{ijt}$ and $\bar{g}_{ijt}$, where $t\in [k-1]$. Concretely, $f_{ijt}$
%Our objective is to transform Equation~\ref{obj} into an linear programming form. By Lemma~\ref{line}, we first replace non-linear functions in Equation~\ref{obj} with linear functions as follows.

\begin{equation}
\label{obj2}
\begin{aligned}
maximize &\sum_{i, j} \sum_{t=1}^{k-1} \bar{g}_{ijt} \cdot x_{ijt} + C \\
s.t. \quad &\sum_{i} \sum_{t=1}^{k-1} \bar{f}_{ijt} \cdot x_{ijt} \geq r_{j} - b_j, \forall j \\
&0\leq x_{ijt}\leq s_{ijt+1}-s_{ijt}, \forall i,j,t\in[1, k-1] \\
&x_{ijt}>0 \implies x_{ijt'}=s_{ijt'+1}-s_{ijt'}, \forall t' < t
\end{aligned}
\end{equation}
Two constant terms, $C=\sum_{i,j} \bar{g}_{ij}(0)$ and $b_j = \sum_{i}\bar{f}_{ij}(0)$, can be ignored in our task. 
We denote the optimized value of the above objective $OBJ_1$. The gap between $OBJ$ and $OBJ_1$ is the caused by the approximation. Without consideration of constraints, the gap between them is
\begin{equation}
\label{eq:err}
\begin{aligned}
    &|OBJ(X)-OBJ_1(X)| \\
    \leq& nm \cdot \sup(|g_{ij}(x)-\bar{g}_{ij}(x)|)\\
    \leq& nm \cdot (|\epsilon_2 + \delta|)= nm\epsilon_2
\end{aligned}
\end{equation}
Here $\delta$ is the difference of $g$ on adjacent separated points, which can be removed when $k$ goes to infinity. On the other hand, feasible solutions of $OBJ$ may not be feasible for $OBJ_1$ and vice versa. The fulfillment of constraints can be similarly proved to have a bound of $n\epsilon_1$ as above. 
% We can use a conservative refinement during the transformation: increase $x_{ij}$ that does not satisfies the constraints. Similar to Equation~\ref{eq:err}, the refinement will change $x_{ij}$ for at most $\epsilon_1 + \delta$, which sequentially enlarge the gap between $OBJ$ and $OBJ_1$ by $nm\epsilon_1 \cdot \sup(|g'|)$.
% On contrary, 
% For a fixed pair of $(i, j)$, $g_{ij}(x)$
% Therefore, $f_{ijt}(x)$ and $g_{ijt}(x)$ are in form of $kx$, where $k$ is the gradient. For convenience, we replace $f_{ijt}(x)$ by $f_{ijt} x$ and replace $g_{ijt}(x)$ by $g_{ijt} x$ (here the values of $f_{ijt}$ and $g_{ijt}$ represent gradients). The last constraint promises the consistency between $f_{ij}$ and $\sum f_{ijt}$ (as well as $g$ and $\sum g_{ijt}$). 
% By Lemma~\ref{convex}, we can remove the last constraint if $f$ and $g$ are convex. To combine these two convex constraint into one, we introduce a equivalent objective with a transformation of $x_{ijt}$:
Next, we are going to eliminate the last constraint of $OBJ_1$. We let $x_{ijt}^{'} = \frac{x_{ijt}}{\bar{f}_{ijt}}$, 
% then the first constraints becomes
% \begin{equation}
% \begin{aligned}
% &\sum_i \sum_{t=1}^{k-1} \bar{f}_{ijt} x_{ijt} \geq r_j - r_b\\
% \implies & \sum_i \sum_{t=1}^{k-1} x_{ijt}^{'} \geq \frac{r_j - r_b}{\bar{f}_{ijt} }.
% \end{aligned}
% \end{equation}
% Then this new constraint will no longer conflict with the last constraint because every assignment can be adjusted to satisfy the last constraint without changing the new constraint. 

\begin{equation}
\label{obj_scaled}
\begin{aligned}
maximize &\sum_{i, j} \sum_{t=1}^{k-1} \frac{\bar{g}_{ijt}}{{\bar{f}_{ijt}}} x'_{ijt} + C  \\
s.t. \quad &\sum_{i} \sum_{t=1}^{k-1} x'_{ijt} \geq \frac{r_{j} + b_j}{\bar{f}_{ijt}}, \forall j \\
&0\leq x'_{ijt}\leq \frac{s_{ijt+1}-s_{ijt}}{\bar{f}_{ijt}}, \forall i,j,t\in[1, k-1] \\
&x'_{ijt}>0 \implies x'_{ijt'}=s_{ijt'+1}-s_{ijt'}, \forall t' < t
\end{aligned}
\end{equation}

If $\frac{\bar{g}_{ijt}}{{\bar{f}_{ijt}}}$ is non-increasing, the last constraint can be eliminated. So the last step of our algorithm is to replace the piece-wise linear functions by the outer convex curve of them. Let $\{(s^c_{ijt}, \frac{\bar{g}^c_{ijt}}{{\bar{f}^c_{ijt}}})\}_{t=1}^{k_c}$ be the outer convex curve of points $\{(s_{ijt}, \frac{\bar{g}_{ijt}}{{\bar{f}_{ijt}}})\}_{t=1}^{k}$. The corresponding programming is

% Now we have removed the conflict between $f_{ijt}$ and Lemma~\ref{convex}. By Lemma~\ref{convexhull}, we can replace $g_{ijt}$ with $g^c_{ijt}$ without loss of the objective. With the replacement, the last constraint can be removed by Lemma~\ref{convex}. After that, the optimal solution is no worse than the one in the original programming. By Lemma~\ref{line}, each point in the convex hull can be reached. Finally, we get the following linear programming that can be solved in the polynomial time:

\begin{equation}
\label{obj_final}
\begin{aligned}
maximize &\sum_{i, j} \sum_{t=1}^{k^c-1} \frac{\bar{g}^c_{ijt}}{{\bar{f}^c_{ijt}}} x'_{ijt}  + C \\
s.t. \quad &\sum_{i} \sum_{t=1}^{k^c-1} x'_{ijt} \geq \frac{r_{j} + b_j}{\bar{f}^c_{ijt}}, \forall j \\
&0\leq x'_{ijt}\leq \frac{s^c_{ijt+1}-s^c_{ijt}}{\bar{f}^c_{ijt}}, \forall i,j,t\in[1, k^c-1] \\
\end{aligned}
\end{equation}
We denote the optimized value of above linear programming is $OBJ_2$ which can be solved in polynomial time. Note that the last constraint in $OBJ_1$ is eliminated here by Lemma~\ref{convex} because $\frac{\bar{g}^c_{ijt}}{{\bar{f}^c_{ijt}}}$ is non-increasing. Also, we then can produce the bonus score assignment by $\{x'_{ijt}\}$ by finding the  $x_{ijt}$ with the maximal $t$ such that $x_{ijt} > 0$ for $(i, j)$ pairs, which can be achieved by the linear combination of its two neighbors as Algorithm~\ref{alg:main} shows. Finally, because of Lemma~\ref{convexhull}, we can see the solution of $OBJ_2$ is no worse than $OBJ_1$ so that satisfies $\mathbb{E}[\hat{f}_{j}] + n\epsilon_1 \geq r_j$  and $E[\hat{g}] + nm\epsilon_2  \geq g^*$. 

%Our final task is to prove the optimal solution for $OBJ_2$ also produces a solution for the original task and satisfies  $\mathbb{E}[\hat{f}_{j}] + n\epsilon_1 \geq r_j$  and $E[\hat{g}] + nm\epsilon_2  \geq g^*$. 

% Lemma~\ref{convexhull} promises $OBJ_2\geq OBJ_1$, so 
% and Lemma~\ref{line} provides the 
%We need one more preparation to produce convex forms of $f_{ijt}$ and $g_{ijt}$. 
\end{proof}

% \begin{coro}
% Given $\{(\bar{f}(x_{1}),\bar{g}(x_{1})), (\bar{f}(x_{2}),\bar{g}(x_{2})), ..., (\bar{f}(x_{k}),\bar{g}(x_{k}))\}$, the protocol in Theorem~\ref{protocol} satisfies $E[\hat{f}(x)]=f(x)$ and $E[\hat{g}(x)]\geq g(x) - \max_{i>1}(\bar{g}(x_{i - 1}) - \bar{g}(x_{i}))$.
% \end{coro}

\subsection{Online test}
We examine our algorithm in the ANONYMOUS platform for a month. The ANONYMOUS platform is one of the largest international online E-Commerce platforms worldwide, and the targeted set of items in this scenario is more than 10\%, which has been a heavy portion of the revenue. 
The baseline method is an industrial-level traffic shaping protocol, which includes a PID module with adjustments from a multi-arm bandit module and has been served our platform for years. Intuitively, PID helps us achieve the exposure requirements and the bandit module can select to push the targeted items to the proper group of users in an online fashion. Each examined method needs to serve more than millions of users per day with support requirements for tens of targeted sets: guarantee a certain count of exposure for each day. We consider three indicators in the online A/B test:
\begin{itemize}
    \item \textbf{Purchase Rate (PR)}. PR is computed as the number of purchases divides by the number of served users.
    \item \textbf{Gross Merchandise Volume (GMV)}. GMV(million dollars) is the total value of the sold items.
    \item \textbf{Compliance Rate (CR)}. The high CR value implies the requirements on exposures of targeted items are better satisfied. Let $C=\{c_1, c_2, ..., c_m\}$ be the actual exposure ratios of $m$ targeted sets, CR is computed as follows:
    \begin{equation}
        \text{CR}(C) = \frac{1}{m}\sum_{i=1}^m \frac{\min(c_i, r_i)}{r_i}.
    \end{equation}
\end{itemize}

\begin{table}[h]
\begin{tabular}{l|lll}
\toprule \hline
 \textbf{Models} & \textbf{PR} & \textbf{GMV} & \textbf{CR} \\ \hline
 No Traffic Shaping & +0.00\% & +0.00\% & 66.14\% \\\hline
 PID + Bandit & -2.53\% & -2.51\% & 81.39\% \\ 
 Ours & -0.89\% & -0.70\% & \textbf{82.65\%} \\
 %Deep LTR & Real-time & $\backslash$ & $\backslash$ \\ \hline
\bottomrule
\end{tabular}
\vspace{0.03em}
\caption{The relative gap on metrics.}\label{tab:online2}
\end{table}

From the above result, we can see traffic shaping has a negative influence on the possibility of purchase, but can expose much more targeted items. Our proposed method can achieve better PR and GMV (greater than 1.5\%) than PID with efficiency selection by the bandit module, which can be translated to a significant improvement that prevents us from losing of millions GMV per day. At the same time, our method can have a compatible CR, which implies that our method can achieve the best trade-off on short-term reward (PR and GMV) and long-term reward (CR) amongst exists methods.

\section{Conclusion}
In this paper, we propose a new framework for traffic shaping in E-Commerce. The proposed framework straightforwardly solves the approximated version of original traffic shaping in expectation, where no accurate conversion rate prediction model needs to be included. The experimental result shows that it can steadily bring revenue to our online system.
\bibliographystyle{ACM-Reference-Format}
\bibliography{sample-base}


\end{document}